%% file: main.tex
\documentclass[10pt,twocolumn,letterpaper]{article}

\usepackage{iccv}
\usepackage{times}
\usepackage{epsfig}
\usepackage{graphicx}
\usepackage{amsmath}
\usepackage{amssymb}
\usepackage{booktabs}
\usepackage{xcolor}
\usepackage{pifont}
\usepackage{multirow}

\usepackage[pagebackref=true,breaklinks=true,letterpaper=true,colorlinks,bookmarks=false]{hyperref}

\newcommand{\method}{\mbox{\textsc{CMOTA}}\xspace}

\usepackage{microtype}
\usepackage{graphicx}
\usepackage{color, colortbl}
\usepackage{booktabs}
\usepackage{wrapfig} 

\usepackage{arydshln}
\usepackage{booktabs}
\usepackage{enumitem}
\usepackage{balance}
\usepackage{combelow}
\usepackage{tabularx}
\usepackage{cite} 
\usepackage[accsupp]{axessibility}  

\usepackage{setspace}

\usepackage[capitalize]{cleveref}
\crefname{section}{Sec.}{Secs.}
\Crefname{section}{Section}{Sections}
\Crefname{table}{Table}{Tables}
\crefname{table}{Tab.}{Tabs.}

\makeatletter
\def\adl@drawiv#1#2#3{%
        \hskip.5\tabcolsep
        \xleaders#3{#2.5\@tempdimb #1{1}#2.5\@tempdimb}%
                #2\z@ plus1fil minus1fil\relax
        \hskip.5\tabcolsep}
\newcommand{\cdashlinelr}[1]{%
  \noalign{\vskip\aboverulesep
           \global\let\@dashdrawstore\adl@draw
           \global\let\adl@draw\adl@drawiv}
  \cdashline{#1}
  \noalign{\global\let\adl@draw\@dashdrawstore
           \vskip\belowrulesep}}
\makeatother

\newcommand\blfootnote[1]{%
  \begingroup
  \renewcommand\thefootnote{}\footnote{#1}%
  \addtocounter{footnote}{-1}%
  \endgroup
}

\iccvfinalcopy 

\def\iccvPaperID{6939} 

\ificcvfinal\fi
\begin{document}

\title{Story Visualization by Online Text Augmentation with Context Memory}

\author{
Daechul Ahn$^{1,\S}$\hspace{1.5em}
Daneul Kim$^{2}$\hspace{1.5em}
Gwangmo Song$^{3}$\hspace{1.5em}
Seung Hwan Kim$^{3}$\\
Honglak Lee$^{3,4}$\hspace{1.5em}
Dongyeop Kang$^{5}$\hspace{1.5em}
Jonghyun Choi$^{1,\dagger}$\vspace{0.3em} \vspace{0.5em}\\
{\hspace{0.5em}$^1$Yonsei University\hspace{0.5em}$^2$GIST\hspace{0.5em}$^3$LG AI Research\hspace{0.5em}$^4$University of Michigan\hspace{0.5em}$^5$University of Minnesota} \vspace{0.5em}\\
{\tt\small \{dcahn,jc\}@yonsei.ac.kr flytodk98@gm.gist.ac.kr gwangmo.song@lgresearch.ai}\\ 
{\tt\small skcruise@gmail.com honglak@eecs.umich.edu dongyeop@umn.edu}
}


\maketitle
\ificcvfinal\thispagestyle{empty}\fi

\input{contents}

\vspace{-1.5em}
{
\footnotesize
\begin{singlespace}
\paragraph{\footnotesize Acknowledgement.}
This work is partly supported by the NRF grant (No.2022R1A2C4002300) 25\%, IITP grants (No.2020-0-01361, AI GS Program (Yonsei University) 5\%, No.2021-0-02068, AI Innovation Hub 5\%, 2022-0-00077 15\%, 2022-0-00113 15\%, 2022-0-00959 15\%, 2022-0-00871 10\%, 2022-0-00951 10\%) funded by the Korea government (MSIT).
\end{singlespace}
}

{\small
\bibliographystyle{ieee_fullname}
\bibliography{egbib}
}

\end{document}

%% file: contents.tex
\begin{abstract}
Story visualization (SV) is a challenging text-to-image generation task for the difficulty of not only rendering visual details from the text descriptions but also encoding a long-term context across multiple sentences. 
While prior efforts mostly focus on generating a semantically relevant image for each sentence, encoding a \emph{context} spread across the given paragraph to generate contextually convincing images (\eg, with a correct character or with a proper background of the scene) remains a challenge. 
To this end, we propose a novel memory architecture for the Bi-directional Transformer framework with an \textit{online text augmentation} that generates multiple 
pseudo-descriptions as supplementary supervision during training for better generalization to the language variation at inference.
In extensive experiments on the two popular SV benchmarks, \ie, the Pororo-SV and Flintstones-SV, the proposed method significantly outperforms the state of the arts in various metrics including FID, character F1, frame accuracy, BLEU-2/3, and R-precision with similar or less computational complexity.\blfootnote{\hspace{-2em}$^\S$: work done while interning at LG AI Research. $^\dagger$: corresponding author. {\bf Code}: \url{https://github.com/yonseivnl/cmota}}
\end{abstract}

\section{Introduction}
\label{sec:intro}

\begin{figure}[t]
    \centering
    \includegraphics[width=1.0\columnwidth] {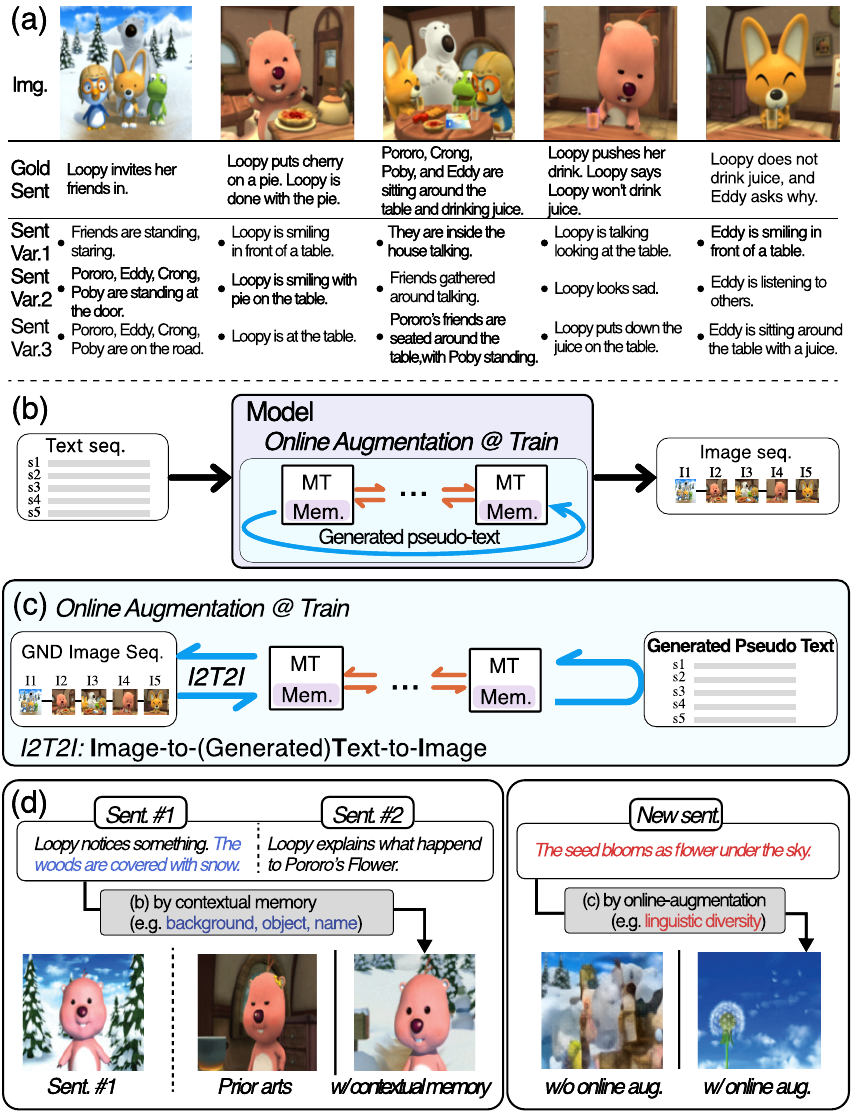}
    \caption{\textbf{Linguistic variations in story visualization and the overview of the proposed method.} 
    (a) An example of a story data with various linguistic variations (Var.\#) for each image. 
    (b) Modeling temporal context spread across sentences by our context memory (Sec.~\ref{sec:method:awm}).
    (c) Addressing linguistic variations by online text augmentation from each image for every epoch (Sec.~\ref{sec:method:cyclic_ptexts}).
    (d) Benefit of the proposed context memory (left) and online text augmentation (right).
    }
    \vspace{-1.5em}
    \label{fig:task}
\end{figure}

Story visualization (SV)~\cite{li2019storygan} is a task of generating a sequence of images from a paragraph, \ie, a sequence of natural language sentences.
It is challenging for the requirement of rendering the visual details in images with convincing background of a scene -- seasonal elements, environmental objects such as table, location, and the proper character appearing, which here we refer to as \emph{context}, spread across the given text sentences.
Specifically, it needs to encode implicit context presented in the given sentences since each one often omit visual details (\ie, they may be spread over the sentences) necessary to generate a semantically correct image.
For example, we can think of Fig.~\ref{fig:task}-(d), where Sent.\#1 (\ie, sentence 1) and Sent.\#2 (\ie, sentence 2) are given as sequences. 
After Sent.\#1 is given, generated image by Sent.\#2 often exhibits a background that is not semantically correct~\cite{li2019storygan,maharana2021improving,maharana2021integrating}.
However, if we use contextual information given in Sent.\#1, ``The woods are covered with snow'', it leads to correct image with matching background.

In addition, widely used benchmark datasets for story visualization task~\cite{Gupta2018ImagineTS,li2019storygan} provide a single text-image pair for training and inference, mostly due to the annotation budget constraint.
This prevents the model from learning language variations, and thus harms the linguistic generalization performance of the model.

To address the aforementioned challenges without requiring large scale data and models, we propose a new memory scheme in bi-directional Transformer for encoding the context which generates pseudo-texts in an online fashion to address linguistic variations at inference.
We call our model as \textbf{C}ontext \textbf{M}emory and \textbf{O}nline \textbf{T}ext \textbf{A}ugmentation or \method for short.
We empirically validate that our model outperforms the state-of-the-art SV methods by large margins on various metrics evaluated with widely used benchmarks in the literature, \ie, Pororo-SV and Flintstones-SV. 

Note that while large pre-trained models \cite{chen2020generative,vaswani2017attention,ding2021cogview,cogview2,make_a_scene,dalle2} have shown great success in synthesizing an image or a video from a language description \cite{tivgan,ramesh2021zero,makeavideo}, huge computational complexity and large training data makes the models prohibited.
Moreover, although we propose and evaluate the model for the standard benchmark datasets without large pretraining data trained with a large model, it would be interesting to apply and evaluate the proposal in large models for further improvement.

We summarize our contributions as follows:
\vspace{-0.5em}
\setlength{\leftmargin}{0.8em}
\begin{itemize}
\setlength\itemsep{-0.3em}
    \item We propose a new memory architecture for Transformer to selectively make use of contexts in a story paragraph.
    \item For better generalization of linguistic variations in the given paragraph at inference, we generate pseudo-texts and augment them in an \emph{online fashion} for richer linguistic supervision. 
    \item Our model significantly outperforms prior arts (even some hyper-scale models) by large margins in five evaluation metrics with similar or less computational complexity.
\end{itemize}

\section{Related Work}
\label{sec:related}

\paragraph{Story visualization.}
StoryGAN~\cite{li2019storygan} is one of the recent methods that utilized a story-level discriminator to improve global consistency in generated images. 
CP-CSV~\cite{song2020character} disentangles figure and background information to enhance character consistency. 
In order to improve global semantic matching between paragraph and generated image sequence, DuCo-StoryGAN~\cite{maharana2021improving} presents a pre-trained video captioner as an auxiliary loss along with other design improvements on top of StoryGAN. 
More recently, VLC-StoryGAN~\cite{maharana2021integrating} utilizes constituency parse-trees and common sense knowledge to improve consistency and an object-level feedback loop to improve image quality. 
Another recent work VP-CSV~\cite{vpcsv} is a two-stage approach, \ie, 1) character generation and 2) background completion, using Transformer model to address this task.
We discuss these in more detail in the supplementary material for the space sake.
But the prior arts largely neglect to encode the story narrative, which is our primary contribution here.

Very recently, Maharana \etal propose a new task setup of story continuation; using first image as a condition.
They fine-tune the large model DALL-E~\cite{ramesh2021zero} for the SV, which they call StoryDALL-E~\cite{storydalle}.
Ours differs from it in several aspects as follows. 
We have explicit memory connections between adjacent image generators using the context memory to globally encode the sentences.
In contrast, \cite{storydalle} utilizes global story embedding as additional input for understanding context.
Further, while it is huge in size (1.3B parameters), trained with 14 million text-image pair, ours are much larger (97M parameters) and outperforms it in multiple metrics by large margins; even in the image quality metric, FID.
Please refer to Sec.~\ref{sec:experiment:quan} for empirical comparisons.

\vspace{-0.5em}
\paragraph{Text-to-video generation.}
Similar to story visualization, text-to-video generation also generates multiple frames from a given text.
Since the pioneering work of Sync-DRAW~\cite{mittal17syncdraw} in text-to-video generation, \cite{Pan2017ToCW, li2018videogen, Gupta2018ImagineTS} utilize generative adversarial network for high-quality image generation.

Recently, high-quality video generation models are proposed, including GODIVA~\cite{godiva} and N{\"U}WA~\cite{Wu2022NWAVS}. 
Furthermore, recent studies generate high-resolution videos, making sequential frames in high-quality~\cite{hong2022cogvideo, makeavideo}.
However, most of state-of-the-art text-to-video model~\cite{hong2022cogvideo,makeavideo} generates a video from a \emph{single} sentence, mostly having consistent backgrounds.
Very recently, there is a method proposed to generate a video from a long paragraph~\cite{penaki}.
Although they generate the video in a long time horizon, they require both a mega-scale model trained with huge data and a detailed paragraph where each sentence is describing the scene that are close in time.
In contrast, story visualization requires generating frames arbitrarily distant in time (\ie, so-called `key-frames') corresponding to different sentences, requiring to generate an image sequence that have contextually convincing background.

\vspace{-0.5em}
\paragraph{Text-to-image generation.}
Text-to-image generation is a sub-problem of story visualization, with literature focusing on semantic relevance and resolution improvements. 
Recently, text-based image synthesis has been greatly improved with the help of a vast amount of training data with a hyper-scale model including DALL-E~\cite{ramesh2021zero} and its successor DALL-E2, CogView~\cite{ding2021cogview} and Make-A-Scene~\cite{makeascene} using a sketch input.

Although text-to-image generation models generates very high-quality images, it may lack encoding the context, metaphoric sentences spread acorss multiple sentences.
In addition, naively using {state-of-the-art} text-to-image generation models is computationally prohibited.
For example, diffusion-based models~\cite{Saharia2022PhotorealisticTD,dalle2} have hyper-scale model size, (\eg, Imagen~\cite{Saharia2022PhotorealisticTD} parameter count of 2-B, DALL-E2~\cite{dalle2} parameter count of 3.5-B) making it non-trivial for applying it in a wide range of inference scenarios that may not have the sufficient computing resource.
Here, we consider relatively light architectures as our base model for computational efficiency.
More discussion are in the supplement.

\vspace{-0.5em}
\paragraph{Transformer using memory.}
To mitigate context fragmentation issue~\cite{dai-etal-2019-transformer_xl}, \ie, losing long-term dependency over a data stream in context, there are efforts to encode long-term contexts in generating an image sequence. 
\cite{dai-etal-2019-transformer_xl,lei-etal-2020-mart} adopt a recurrent path into transformer architecture.
Specifically, the modeling of new data segments is conditioned on historical hidden states produced in the previous time step and uses highly summarized memory states.
Unlike the conventional memory architectures, we propose a novel memory that has a dense connection from the past with attentive weighting schemes for their better usage (Sec.~\ref{sec:method:awm}). 

\vspace{-0.5em}
\paragraph{Online augmentation.}
While offline data augmentation that prepares data outside of learning process is prevalent in many literature~\cite{info11020125,Jackson2019StyleAD,Lim2018DOPINGGD}, online-augmentation that depends on training is seldom explored especially in story visualization task~\cite{tang2020onlineaugment}.
\cite{tang2020onlineaugment,vqa_late} propose an online augmentation for image classification and VQA task.
\cite{mounsaveng2020learning} uses bi-level optimization in image classification.
In medical domain, \cite{oada21} investigate the online augmentation for personalized image based Histopathology diagnosis~\cite{oada21}.
Note that we show the benefit of online augmentation over the offline augmentation in the SV for the first time in this literature.

\section{Approach}

There are multiple challenges in story visualization, including (1) generating semantically natural images without artifacts from a description, (2) encoding  the \emph{context} (\eg, consistent background of a scene - seasonal elements, environmental objects such as table and location, and consistent characters) spread across the sentences in a given paragraph~\cite{li2019storygan} and (3) addressing the linguistic variation at the inference time (\ie, a given text description may not be in different writing style).

For photo-realistic image generation from a language description, we recently witness unprecedented improvements in the quality of generated images by the help of large scale model~\cite{chen2020generative,vaswani2017attention,ding2021cogview,cogview2,make_a_scene,dalle2}.
To leverage the large model's benefits for the story visualization task, we may use it to generate an image sequence by \emph{gradually} concatenating sentences to visualize the story (\ie, use the first sentence to generate the first image, use a concatenated sentence of the first and the second to generate the second, and so on).
Although this may generate a single photo-realistic image, it is not able to capture the context sparsely spread across the sentences as story progresses (see supplement for more discussion).

A more involved way of using the large model for SV is to fine-tune it for the downstream SV or using a video generation models.
However, as the most high-performing models are huge in size and the codes and the pre-trained models are not publicly available, the computational cost and reproducibility reduces practicality.

\label{sec:method}
\begin{figure}[t]
    \centering
    \includegraphics[width=1.0\columnwidth] {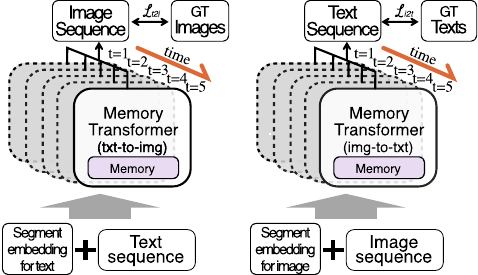}
    \caption{\textbf{Bi-directional (between text and image modality) Transformer for text or image `sequence' generation with the proposed memory.} 
    Our bi-directional `base' model can simultaneously generates a sequence of both text and image from $t=1...5$ with the proposed memory (Sec.~\ref{sec:method:awm}). 
    We use a segment embedding to indicate input modality and enable output of any modality.}
    \label{fig:overview}
\end{figure}

For a given dataset without using large extra training data, we propose a new memory module to better encode past information with a Transformer by an attentively weighted densely connected architecture (Sec.~\ref{sec:method:awm}). 
We further propose to generate pseudo-texts during the learning process to augment (online-augmentation) for better linguistic generalization without requiring large external data by learning the \emph{bi-directional} Transformer~\cite{kim2021lverse,qiao2019learn,reed2016generative} in both directions of generating images from texts and \emph{vice versa} as illustrated in Fig.~\ref{fig:overview} (Sec.~\ref{sec:method:cyclic_ptexts}).

\vspace{-0.5em}
\paragraph{Base model.}
As shown in Fig.~\ref{fig:overview}, we use bi-directional (\ie, multi-modal) transformer that iteratively generates images and texts in both ways. 
Similar to~\cite{ramesh2021zero}, the image tokens are sequentially predicted from the input text sequences by the Transformer.
Then, the decoder of the VQ-VAE~\cite{vqvae} translates the predicted image tokens into image sequence.
The text tokens are also sequentially predicted from the input image token sequence by the same Transformer.  

Particularly, for the \emph{bi-directional multi-modal} generation, \ie, generating simultaneously text and image from the unified architecture, we add two embeddings; a positional embedding for absolute position between tokens and a segment embedding for distinguishing source and target.
Tokens of a text (\{$t_{1}, ..., t_{m}$\}) and an image (\{$z_{1}, ..., z_{n}$\}) ($m,n$: \# of tokens for text and image) are fed into the Transformer to predict tokens in the other modality in multiple epochs with the following objective function written as:
\begin{equation}
\begin{split}
    \mathcal{L}_{j, t2i} &= \sum_{k=1}^n -\ln p_j(z_k | t_1, ..., t_m, z_1, ..., z_{k-1}), \\
    \mathcal{L}_{j, i2t} &= \sum_{k=1}^m -\ln p_j(t_k | z_1, ..., z_n, t_1, ..., t_{k-1}), \\
    \mathcal{L}_j &= \mathcal{L}_{j,t2i} + \lambda_{1} \mathcal{L}_{j, i2t},
\end{split}
\label{eq:bi_loss}
\end{equation}
where $p_j(\cdot)$ is a likelihood of $j$-th generated tokens in one modality given the other modality, $L_{j,t2i}$ is a loss corresponding to $j$-th text-to-image generation (\ie, negative log likelihood), $L_{j,i2t}$ is vice versa. 
$\lambda_{1}$ is a balancing hyper-parameter.

To train the model, we iteratively train the generation model in each modality multiple times per each epoch.

\begin{figure}[t]
    \centering
    \includegraphics[width=1.0 \linewidth]{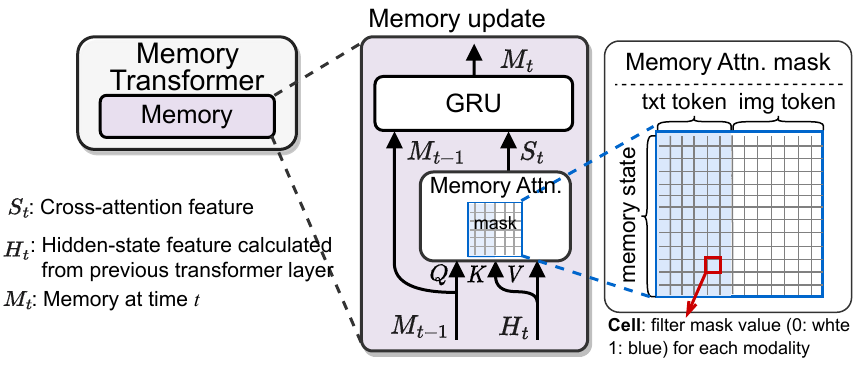}
    \caption{\textbf{Proposed memory module for the bi-directional Transformer.} Memory module updates current memory $M_{t}$ by utilizing memory attention and GRU operation with propagated memory, $M_{t-1}$, and previously calculated hidden state, $H^{l}_{t}$. To produce image, we propose a novel memory attention (Attn.) mask denoted by blue squares to select only text tokens for less bias to previously generated image tokens. Within the memory attention mask, white-colored sqaures filter out the content, \ie, image (img) token, whereas the blue squares allow the text (txt) token to propagate through for memory interaction.}
    \label{fig:method}
\end{figure}

\subsection{Context Memory}
\label{sec:method:awm}
To encode the context and calculate the propagated memory, as depicted in Fig.~\ref{fig:method}, we first apply cross attention between the current hidden state, $H^{l}_{t} \in \mathcal{R}^{T_c \times d}$ calculated from $(l-1)$-th transformer layer and memory state at time $(t-1)$, $M_{t - 1} \in \mathcal{R}^{T_M \times d}$ ($T_c$: \# tokens, $T_M$: \# memory states, $d$: hidden state dimension).
We then obtain $S_t = Attn(M_{t-1}, H_t, H_t)$, where $Attn(Q, K, V) := Softmax \left( {\frac{QK^T}{\sqrt{d_q}}} \right) {V}$ and $d_q$ is a query dimension ($Q$), with the memory attention mask depicted in the blue box in Fig.~\ref{fig:method}.

In particular, we apply a memory mask in the attention operation to select text tokens as memory content (depicted as blue-shaded grid cells in the right of Fig.~\ref{fig:method}) because including image content as memory could be strong constraint for text-to-image generation (see empirical studies of the proposed mask in the supplementary material).
Then we feed the $S_t$ and $M_{t-1}$ to the GRU in order to compute the information to be stored in the memory, $M_{t}$, propagated to ($t+1$). 

\vspace{-0.5em}
\paragraph{Memory connection.}
\begin{figure}[t]
    \centering
    \includegraphics[width=1.0\linewidth]{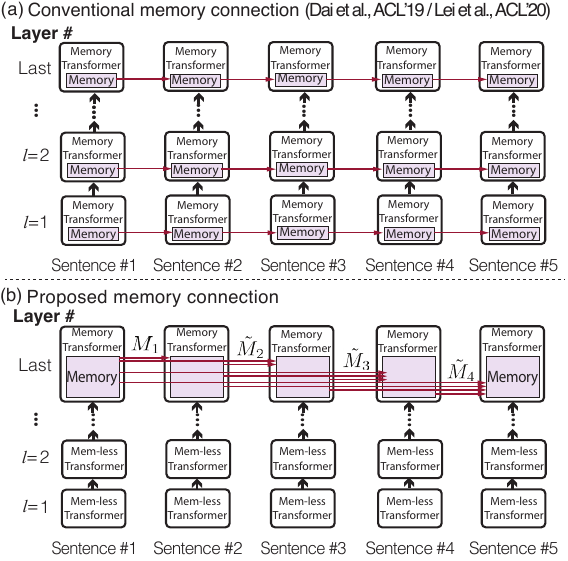}
    \caption{\textbf{Comparison of our memory connection scheme to the conventional one~\cite{lei-etal-2020-mart,dai-etal-2019-transformer_xl}.} We depict the memory connections in the multi-layer Transformer architecture per each sentence input (unfolded in a row). Unlike (a) the previous memory architecture having a single memory path from the immediate past in all layers, (b) ours has a dense connection from all the pasts to be attentively weighted (denoted here by $\bar{M}$'s) in only the last layer. We discuss its empirical benefit in Sec.~\ref{sec:ablation}.}
    \label{fig:method:mem_arch}
\end{figure}

Following~\cite{lei-etal-2020-mart,dai-etal-2019-transformer_xl}, we stack multiple Transformer's layers, depicted in Fig.~\ref{fig:method:mem_arch}-(b).
Compared to conventional memory updating architectures with serial hidden state connections in all layers (Fig.~\ref{fig:method:mem_arch}-(a)), our approach only has memory connections in the last layer, which we call it as partial-level memory augmented (PMA). 
This is because the later layers (closer to the last layer) achieve better representations with higher-level features, abstract and structured representations~\cite{park2019relational,pkt_eccv,tian2019crd} with improved computational efficiency, which is unlike prior works (see Sec.~\ref{sec:ablation} and supp for further discussion).

Additionally, not all historic information is equally important for generating an image at a time step.
Similar to the masked self-attention \cite{vaswani2017attention}, we attentively weight the past information for better modeling of sparse context as:
\begin{equation}
    \begin{split}
        \bar{M}_{1:(t-1)} &= Attn(M_{(t-1)}, M_{[1:(t-2)]}, M_{[1:(t-2)]}), \\
        \tilde{M}_{(t-1)} &= [M_{t-1}; \bar{M}_{1:(t-1)}], ~~~~ (3 \leq t \leq 5),\\
        H^{l}_{t} &= Attn(H^{l}_{t}, [H^{l}_{t}; \tilde{M}_{(t-1)}], [H^{l}_{t}; \tilde{M}_{(t-1)}]),
    \end{split}
    \label{eq:attention}
\end{equation}
where $M_{(t-1)}$ is a memory at time $(t-1)$ and $[1:(t-2)]$ refers to the concatenation from time 1 to $(t-2)$ as depicted in Fig.~\ref{fig:method:mem_arch}-(b).
By doing so, we fuse the contextual information into current hidden state $H_t$.
At $t=2$, we use $M_{1}$, instead of $\tilde{M}_{1}$ as the $M_{1}$ is only available.
We call it as attentively weighted memory (AWM).

\subsection{Online Text-Augmentation}
\label{sec:method:cyclic_ptexts}
Vedamtam \etal argue that multiple descriptions for an image help generalization as they address language variations presented in descriptions~\cite{Vedantam_2015_CVPR}. 
Hence, a number of image captioning datasets \cite{Gurari_2019_CVPR,coco_data,sidorov2019textcaps} provide multiple natural language directives, obtained by multiple human annotators, in both training and evaluation splits. 
But the SV benchmark datasets~\cite{li2019storygan} provide only a single sentence per an image.

\begin{figure}[t]
    \centering
    \includegraphics[width=1.0\columnwidth]{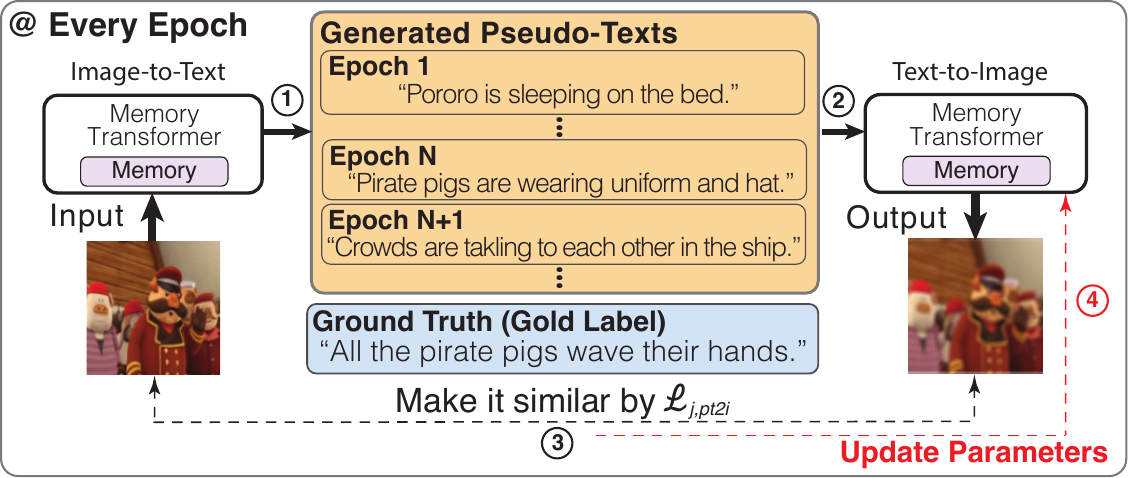}
    \caption{\textbf{Online text augmentation.} We iteratively augment the pseudo-text for better linguistic generalization.
    (1) generating a pseudo-text at each epoch for each ground-truth image, without gradient flow. 
    (2) generating an image for every epoch. 
    (3,4) learn the text-to-image model with $L_{CE}$. 
    In early epochs, it generates less meaningful pseudo-text (thus discard them), but produces pseudo-texts matching with the input image as the training progresses.
    }
    \vspace{-1em}
    \label{fig:bi_direction}
\end{figure}

To address the linguistic variations of a text input at inference process, we first consider to generate a pseudo text by a well-trained image-to-text generation model, which we refer it as \emph{offline-augmentation}~\cite{li2022blip}.
But the offline augmentation generates only a single sentence, which may not provide sufficient diversity. 
Instead, we propose to generate multiple pseudo-texts and augment them in an \emph{online} fashion when training our model to increase the diversity. 
We call it \emph{online text augmentation}, depicted in Fig.~\ref{fig:bi_direction}.

\begin{figure*}[t]
    \centering
    \includegraphics[width=0.9\textwidth]{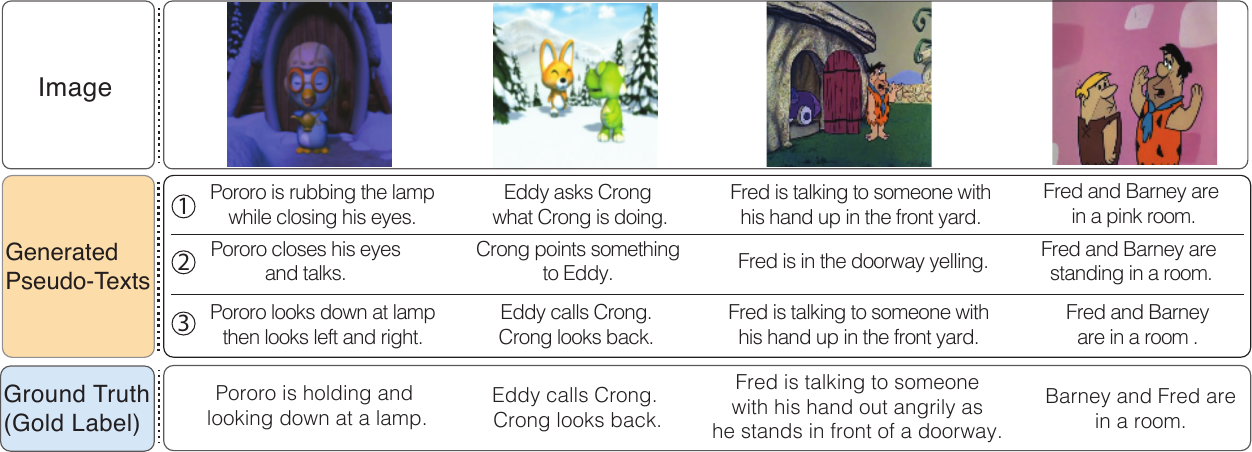}
    \caption{\textbf{Generated pseudo-texts by the proposed method during training.} We depict the generated pseudo-texts in training procedure. We utilize them as supplementary supervisions to address linguistic diversity of a sentence at inference.}
    \vspace{-1em}
    \label{fig:contextual_info}
\end{figure*}

Thanks to our {bi-directional multi-modal} architecture, we can naturally integrate the process of generating pseudo-texts to the process of learning image-to-text model and the text-to-image generation model as depicted in Fig.~\ref{fig:bi_direction}. 
In the early epochs, less meaningful sentences are generated, but as training progresses, more meaningful sentences are generated (see orange box in Fig.~\ref{fig:bi_direction}). 
As a side-product, by supervising the model learning with intermediate goals at each time step, we expect to expedite the convergence of learning. 
When we use the online text augmentation, we can rewrite the objective as:
\begin{equation}
\begin{split}
    \mathcal{L}_{j, pt2i} &= \sum_{k=1}^n -\ln p_j(z_k | \hat{t}_1, ..., \hat{t}_m, z_1, ..., z_{k-1}), \\
    \mathcal{L}_j &= \mathcal{L}_{j,t2i} + \lambda_{1} \mathcal{L}_{j, i2t} + \lambda_{2} \mathcal{L}_{j, pt2i},
\end{split}
\label{eq:total_loss}
\end{equation}
where $\mathcal{L}_{j, pt2i}$ is the additional loss with the augmented pseudo-texts and $\mathcal{L}_{j, t2i}, \mathcal{L}_{j, i2t}$ are defined in Eq.~\ref{eq:bi_loss} and $\lambda_{1}$ and $\lambda_{2}$ are balancing hyper-parameters.
$\hat{t}$ means the pseudo-text token, predicted during online augmentation without gradient flow.
Fig.~\ref{fig:contextual_info} shows pseudo-texts generated by our method.
Detailed training procedure and more examples are in the supplementary material.

\section{Experiments}

\subsection{Experimental Setup}

\paragraph{Datasets.} We use two popular benchmark datasets for evaluating the task of story visualization.
Following~\cite{maharana2021improving}, we use {Pororo-SV} dataset for story visualization task without data overlap. 
Following prior work~\cite{maharana2021integrating}, we use {Flintstones-SV} which was originally exploited in the text-to-video synthesis task~\cite{Gupta2018ImagineTS}.

\vspace{-0.5em}
\paragraph{Metrics.} For the evaluation metrics, we use {Fr$\acute{\text{e}}$chet Inception Distance (FID)} (used in \cite{song2020character,maharana2021integrating}) for visual quality of the generated images, character classification score (Char. F1, Frm. Acc.)~\cite{li2019storygan} for character consistency, and global semantic matching (B-2/3, R-Prec.)~\cite{maharana2021improving,maharana2021integrating} between generated image sequence and story paragraph..
We elaborate on the details of the datasets and implementation in the supplementary material for the sake of space.

\vspace{-0.5em}
\paragraph{Baselines.} We compare our method to a number of prior arts including state-of-the-art story visualization methods such as StoryGAN~\cite{li2019storygan}, CPCSV~\cite{song2020character}, DuCo-StoryGAN~\cite{maharana2021improving}, VLC-StoryGAN~\cite{maharana2021integrating} and VP-CSV~\cite{vpcsv} for both quantitative and qualitative analysis in the story visualization setting.
Moreover, we compare~\cite{storydalle} in the story continuation setting. 

\vspace{-0.5em}
\paragraph{Implementation details.} To train our model, we first tokenize the text and image inputs.
Particularly, the image inputs are encoded by the encoder of the VQ-VAE~\cite{vqvae} into image tokens and decoded by the decoder of the VQ-VAE for generating image sequence.
Once the inputs are tokenized, we append two special embedding tokens to the text and image tokens. These tokens signify the beginning of each modality, denoted as `SOS' (start of sentence) and `SOI' (start of image).
Subsequently, we add two embeddings: (1) a positional embedding to indicate absolute position of each token and (2) a segment embedding to distinguish source and target modality, \eg, for text-to-image generation task, the source corresponds to the text while the target corresponds to the image.
Through the utilization of segment embeddings to distinguish between source and target, we enable the model to generate target data from source data within a unified architecture, as illustrated in Fig.~\ref{fig:overview}.
Here, we set the token length for each modality as $T_{text} = 80$ and $T_{image} = 256 \ (\ie, 16\times16)$, the hidden dimension size to 512, the number of transformer layer to 6, and the number of attention heads to 16, thereafter the number of trainable parameters approximately 93.7-M as shown in Tab.~\ref{tab:main_quan}.
Particularly, we generate an image per each sentence without any sampling process.
We train the \method using AdamW~\cite{adamw} optimizer with $\beta_{1}=0.9$, $\beta_{2}=0.95$, $\epsilon=1e-8$, weight decay multiplier with $1e-2$ and learning rate with $4.5e-6$ multiplied by batch size.
Moreover, we set the hyper-parameters for loss balancing in Eq.~\ref{eq:total_loss} as $\lambda_{1}=1.0$ and $\lambda_{2}=0.5$ and the number of memory state, $T_{M}$ as 1. 
Finally, to generate a high-resolution image, we modified the VQ-VAE with the additional trainable parameters (\ie, $7.1$-M shown in Tab.~\ref{tab:main_quan}).
This modification allows for the generation of high-resolution images, increasing the image resolution from $64\times64$ to $128\times128$.

\begin{table*}[t]
\centering
\resizebox{0.95\linewidth}{!}{
    \begin{tabular}{cclcccccc}
    \toprule
    Dataset & Resolution &Methods & \# Params. & FID$\downarrow$ & Char. F1$\uparrow$ & Frm. Acc.$\uparrow$ & BLEU-2/3$\uparrow$ & R-Prec.$\uparrow$\\
    \midrule
    && StoryGAN~\cite{li2019storygan}                    & - & 158.06 & 18.59 & 9.34 & 3.24 / 1.22 & 1.51 $\pm$ 0.15\\
    &&CP-CSV~\cite{song2020character}                & -  & 140.24 & 21.78 & 10.03 & 3.25 / 1.22 & 1.76 $\pm$ 0.04\\
    &&DuCo-StoryGAN~\cite{maharana2021improving}     & 101M & 96.51 & 38.01 & 13.97 & 3.68 / 1.34 & 3.56 $\pm$ 0.04\\
    &&VLC-StoryGAN~\cite{maharana2021integrating}    & 100M & 84.96 & 43.02 & 17.36 & 3.80 / 1.44 & 3.28 $\pm$ 0.00\\
    &&VP-CSV~\cite{vpcsv}                            & - & 65.51 & \textbf{56.84} & \textbf{25.87} & 4.45 / 1.80 & 6.95 $\pm$ 0.00 \\
    \rowcolor[RGB]{230,230,230}
    \cellcolor{white}&\cellcolor{white}\multirow{-6}{*}{\rotatebox[origin=c]{0}{$64\times64$}}&\method (Ours)                & 96.6M  & \textbf{52.13} & 53.25 & 24.72 & \textbf{4.58 / 1.90} & \textbf{7.34 $\pm$ 0.03}\\
    \cmidrule{2-9}
    &&VLC-StoryGAN-HR$^{\dagger}$~\cite{maharana2021integrating}  & 102.6M & 97.08 & 40.36 & 17.17 & 3.89 / 1.58 & 3.47 $\pm$ 0.03\\
    \rowcolor[RGB]{230,230,230}
    \cellcolor{white}\multirow{-9}{*}{\rotatebox[origin=c]{90}{Pororo-SV}}
    &\cellcolor{white}\multirow{-2}{*}{\rotatebox[origin=c]{0}{$128\times128$}}&\method-HR (Ours)               & 103.7M & \textbf{52.77} & \textbf{58.86} & \textbf{28.89} & \textbf{5.45 / 2.34} & \textbf{16.36 $\pm$ 0.05}\\
    \midrule
    &&StoryGAN~\cite{li2019storygan}                 & -    & 127.19 & 46.20 & 32.96 & 13.87 / 7.83 & 1.72 $\pm$ 0.18\\
    &&DuCo-StoryGAN~\cite{maharana2021improving}     & 101M & 78.02 & 54.92 & 36.34 & 15.48 / 9.17 & 2.64 $\pm$ 0.17\\
    &&VLC-StoryGAN$^{\ddag}$~\cite{maharana2021integrating}  & 100M &72.87 &58.81 &39.18 & - & -\\
    \rowcolor[RGB]{230,230,230}
    \cellcolor{white}&\cellcolor{white}\multirow{-4}{*}{\rotatebox[origin=c]{0}{$64\times64$}}&\method (Ours)                    & 96.6M   & \textbf{36.71} & \textbf{79.74} & \textbf{66.01} & \textbf{19.85} / \textbf{12.98} & \textbf{10.50} $\pm$ \textbf{0.35}\\
    \cmidrule{2-9}
    \rowcolor[RGB]{230,230,230}
   \cellcolor{white}\multirow{-5.5}{*}{\rotatebox[origin=c]{90}{Flintstones-SV}}
   &\cellcolor{white}$128\times128$&\method-HR (Ours)           
                                                        & 103.7M    & 54.81 & 86.44 & 74.06 & 22.18 / 15.17 & 25.71 $\pm$ 0.70\\
    \bottomrule
    \hline
    \end{tabular}
}
    \vspace{0.5em}
    \caption{\textbf{Quantitative comparison with the state of the arts.} On the test split of Pororo-SV and Flintstones-SV. $\#$ Params. refers to the number of trainable parameters. Char. F1 refers to character F1 score. Frm. Acc. refers to frame accuracy. R-Prec. refers to R-Precision. $\downarrow$ indicates `lower the better' and $\uparrow$ indicates `higher the better'. Experiments are done in both $64\times64$ and $128\times128$ resolutions. `HR' refers to its high resolution ($128\times128$) version. $\dagger$ indicates our results with author's implementation ({\footnotesize\url{https://github.com/adymaharana/VLCStoryGan}}). $^\ddag$ indicates the absolute values that we computed by the given relative values in~\cite{maharana2021integrating}.}
    \vspace{-0.5em}
\label{tab:main_quan}
\end{table*}

\subsection{Quantitative Analysis}
\label{sec:experiment:quan}

\subsubsection{Comparison with the State of the Arts}

In Table~\ref{tab:main_quan}, we summarize the performance of the prior arts and the proposed \method in Pororo-SV~\cite{li2019storygan} and Flintstones-SV~\cite{Gupta2018ImagineTS} dataset in the various metrics. 
By default, all methods generate $64\times64$ images along with the higher resolution $128\times128$ images. 
\method outperforms existing methods by a large margin in both benchmarks in most of metrics.

The very recent method VP-CSV~\cite{vpcsv} performs \emph{on par} to our~\method while ours still outperforms in FID, BLEU and R-precision, implying that \method generates high quality images maintaining global semantic matching between story paragraph and images better than the VP-CSV.
The lower performance than the VP-CSV in Char. F1 score and Frm. Acc. is attributed to the specialized character-centric module that only focuses the model to generate accurate characters for each image at the expense of the other performance metrics.
In addition, our \method-HR model even outperforms the VP-CSV~\cite{vpcsv} by large margin without the specialized character-centric module. 

\vspace{-0.5em}
\paragraph{Low resolution to high resolution.}
Not surprisingly, \method-HR outperforms \method with default (low-) resolution (\ie, $64\times64$) in almost all metrics.
Interestingly, however, VLC-StoryGAN-HR~\cite{maharana2021integrating} shows similar performance to $64\times64$ model in Frame accuracy, BLEU, R-precision, and drops performance in all other metrics.
For the increased FID scores by the VLC-StoryGAN when goes to high resolution, we hypothesize that the constituency parse trees of input sentences in the VLC-StoryGAN increases the difficulty in generating visual details as the images contains more visual details in high resolution. 

In addition, we observe that the performance drop of FID in both our~\method and VLC, when goes to high resolution.
It is because as high-resolution image requires the model to depict fine-grained details, generation task would be more difficult compared to the low-resolution image generation.
In particular, the drop of FID in Flintstones-SV is larger than that in Pororo-SV. 
It is because there are more complicated visual details in the Flintstones-SV than the Pororo-SV dataset.
Note that the FID drops by our method from low-resolution to high resolution is less than the VLC in Pororo-SV.
Furthermore, in terms of computational cost, our \method gains performance boost by using high-resolution effectively with just adding relatively small computational cost on top of original $64\times64$ model.

\begin{figure}[t] 
    \centering
    \includegraphics[width=1.0\columnwidth]{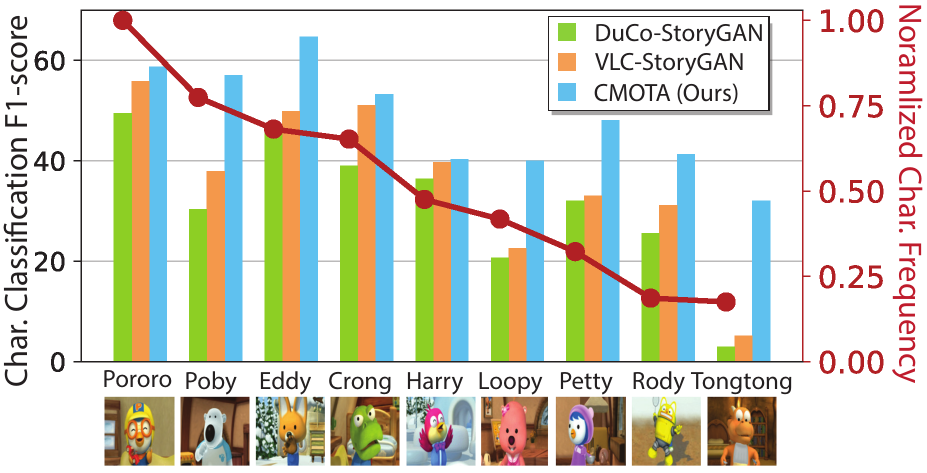}
    \caption{\textbf{Per-character Classification F1-score on the Test Split of Pororo-SV Dataset.} Normalized character (Char.) frequency is shown in red line. It represents the relative number of characters compared to the count of Pororo in training dataset. While DuCo/VLC-StoryGAN shows degraded performance as the character frequency decreases, \method outperforms them across all the characters.}
    \vspace{-1em}
    \label{fig:char_f1}
\end{figure}

\vspace{-0.5em}
\paragraph{Character consistency.}
Although our method does not explicitly enforce the character semantic preservation, the proposed memory delivers the semantic across the sentence implicitly. 
We here investigate the character context preservation performance by comparing our method to~\cite{maharana2021improving, maharana2021integrating} on per-character classification F1-score on test split of Pororo-SV dataset and normalized character frequency in train split dataset as illustrated in Fig.~\ref{fig:char_f1}.
Especially, it is interesting to compare with \cite{vpcsv} as it is explicitly proposed for that purpose.

As shown in the figure, our method outperforms prior arts~\cite{maharana2021improving,maharana2021integrating} by large margins.
Although we did not have any special module for character information preserving, unlike~\cite{vpcsv}, our model maintains such information over multiple sentences, even outperforming the image quality denoted as FID by $-13.38$.

\subsubsection{Ablation Studies}
\label{sec:ablation}

\begin{table*}[t]
\begin{center}
\resizebox{0.9\linewidth}{!}{
    \begin{tabular}{lcccccc}
    \toprule
    Methods & \# Params. & FID$\downarrow$ & Char. F1$\uparrow$ & Frm. Acc.$\uparrow$ & BLEU-2/3$\uparrow$ & R-Prec.$\uparrow$\\
    \midrule
    VLC-StoryGAN~\cite{maharana2021integrating}       & 100M  & 84.96 & 43.02 & 17.36 & 3.28 / 1.44 & 3.28 \\
    \midrule
    Single Directional Transformer (Tr.)                & 93.7M & 63.88 & 45.48 & 18.44 & 4.18 / 1.69 & 5.67\\
    + Partial-level Memory Connection (PMA)        & 95.8M & 59.05 & 49.72 & 21.79 & 4.41 / 1.77 & 6.28\\
    ~~+ Attentional Weighted Memory (AWM)       & 96.6M & 56.98 & 50.15 & 22.01 & 4.50 / 1.91 & 6.79\\
    ~~~~+ Bi-directional Training (Bi.)           & 96.6M & 54.69 & 51.27 & 22.15 & 4.52 / 1.84 & 7.12\\
    ~~~~~~+ Online Augmentation (Full model)                      
                                                    & 96.6M    & \textbf{52.13} & \textbf{53.25} & \textbf{24.72} & \textbf{4.58} / \textbf{1.90} & \textbf{7.34}\\
    \bottomrule
    \end{tabular}
    }
    \vspace{0.5em}
    \caption{\textbf{Benefit of the proposed components.} On the test split of Pororo-SV Dataset. As a baseline, we use a single directional transformer-based text-to-image generation model. We gradually add each of proposed components to investigate its effect for improving performance, \ie, partial-level memory connection (Sec.~\ref{sec:method:awm}), attentively weighted memory (Sec.~\ref{sec:method:awm}), bi-directional training (Sec~\ref{sec:method:cyclic_ptexts}), and online-augmentation (Sec.~\ref{sec:method:cyclic_ptexts}).}
    \vspace{-1.5em}
\label{tab:ablation_test}
\end{center}
\end{table*}

\paragraph{Benefit of the proposed architectural components.} To investigate the benefit of each proposed architectural component, we build up the full model with the components from the vanilla, single-directional transformer without memory as summarized in Tab.~\ref{tab:ablation_test}.

Using the proposed partial-level connected memory architecture only with the vanilla transformer (Tr.), the generation performance improves in all metrics compared to the Tr.
Also, if we compare it with all-level connected memory architecture~\cite{lei-etal-2020-mart}, PMA shows better performance in all metrics as in Tab.~\ref{tab:memory_design_comp} (see supplementary material for more experimental analysis for various memory connection schemes).
Employing attentional weighted memory (AWM) improves performance overall with slight degradation in FID ($+0.12$). 
The bi-directional learning scheme (Bi.) improves performance in all metrics except the character classification related metrics, \ie, Char.F1 and Frm. Acc.
It is because there are captions that does not contain characters' name, just mentioning characters as `friends', making the model difficult to generate specific characters.

\vspace{-0.5em}
\paragraph{Benefit of the text augmentation.} By augmenting the generated text in offline manner (\ie, using an image captioning model pretrained on the dataset of interest), we observe overall performance increase in all metrics.
We hypothesize its reason as that the pseudo-text may convey contextual information such as characters' existence or description of background, thereby making the generation model to be robust to language variation as shown in a first row of Tab.~\ref{tab:ablation_test_online_offline}.
By the online augmentation of pseudo texts (\ie, gradually updating text generator, thus generating a diverse and gradually varying set of pseudo-texts), the performance further improves, as shown in Tab.~\ref{tab:ablation_test} and Tab.~\ref{tab:ablation_test_online_offline}.

\begin{table}[t]
\begin{center}
\resizebox{1.0\linewidth}{!}{
    \begin{tabular}{lcccccc}
    \toprule
    Memory Architecture & \# Params. & FID$\downarrow$ & Char. F1$\uparrow$ & Frm. Acc.$\uparrow$ & BLEU-2/3$\uparrow$ & R-Prec.$\uparrow$\\
    \midrule
    All-Level Connection~\cite{lei-etal-2020-mart}            & 118M  & 61.23 & 47.21 & 19.21 & 4.21 / 1.72 & 6.08\\
    Partial-level Connection (Ours)  & 95.8M & 59.05 & 49.72 & 21.79 & 4.41 / 1.77 & 6.28\\
    \bottomrule
    \end{tabular}
    }
    \vspace{0.5em}
    \caption{\textbf{Benefit of the proposed memory connection scheme.} Proposed scheme outperforms conventional memory connection that uses all-level connections~\cite{lei-etal-2020-mart} (Fig.~\ref{fig:method:mem_arch}-(a)) with less number of parameters (test split of Pororo-SV).}
    \vspace{-1.5em}
    \label{tab:memory_design_comp}
\end{center}
\end{table}

\begin{table}[t]
\begin{center}
\resizebox{1\linewidth}{!}{
    \begin{tabular}{lcccccc}
    \toprule
    Augmentation & \# Params. & FID$\downarrow$ & Char. F1$\uparrow$ & Frm. Acc.$\uparrow$ & BLEU-2/3$\uparrow$ & R-Prec.$\uparrow$\\
    \midrule
    Offline     & 96.6M & 54.51 & 51.32 & 22.31 & 4.50 / 1.90 & 7.09\\
    Online (Ours)                      
                            & 96.6M    & \textbf{52.13} & \textbf{53.25} & \textbf{24.72} & \textbf{4.58} / \textbf{1.90} & \textbf{7.34}\\
    \bottomrule
    \end{tabular}
    }
    \vspace{0.5em}
    \caption{\textbf{Benefit of the online augmentation.} On the Pororo-SV test split. The proposed online augmentation outperforms the offline augmentation using a pretrained captioner~\cite{lei-etal-2020-mart}.}
    \vspace{-1.5em}
    \label{tab:ablation_test_online_offline}
\end{center}
\end{table}

\subsubsection{Comparison with Large-Scale Models}
\label{sec:exp:largescale}
Despite the unfairness of comparison to the large-scale models due to the model size, we compare \method to larger models~\cite{storydalle} in Tab.~\ref{tab:storycont_quan}. 
it is surprising to see that \method outperforms on an evaluation metric of character classification (\ie, Char.F1 and Frm. Acc.), regardless of using prompt-tuning or fine-tuning of StoryDALL-E.

But as expected, the image quality denoted as FID is worse than the StoryDALL-E thanks to the huge size of the model and training dataset.
Nevertheless, FID score is slightly better (Pororo-SV) and on-par (Flintstones-SV) compared to `prompt-tune' that updates 30\% of parameters~\cite{storydalle}.
We discuss more about this in the supplementary material.

\begin{table}[t]
\centering
\resizebox{1\linewidth}{!}{
    \begin{tabular}{cllccc} 
    \toprule
    & Methods & \# Params. & FID$\downarrow$ & Char. F1$\uparrow$ & Frm. Acc.$\uparrow$\\
    \midrule
    &StoryDALL-E (prompt-tune)                &1.3B     & 61.23 & 29.68 & 11.65\\
    &StoryDALL-E (fine-tune)                 &1.3B & 25.90 & 36.97 & 17.26\\
    &MEGA-StoryDALL-E (fine-tune)            &2.8B & \textbf{23.48} & 39.91 & 18.01\\
    \cmidrule{2-6}
    \rowcolor[RGB]{230,230,230}
    \cellcolor{white}\small\multirow{-4.5}{*}{\rotatebox[origin=c]{90}{Pororo-SV}}&\method (Ours)                  
                                            & \textbf{96.6M}     & 55.26 & \textbf{51.48} & \textbf{22.73}\\
    \midrule
    &StoryDALL-E (prompt-tune)                 &1.3B    & 53.71 & 42.38 & 32.54\\
    &StoryDALL-E (fine-tune)                &1.3B   & 26.49 & 73.43 & 55.19\\
    &MEGA-StoryDALL-E (fine-tune)           &2.8B   & \textbf{23.58} & 74.26 & 54.68\\
    \cmidrule{2-6}
    \rowcolor[RGB]{230,230,230}
    \cellcolor{white}\small\multirow{-4.5}{*}{\rotatebox[origin=c]{90}{Flintstones-SV}}&\method (Ours)             & \textbf{96.6M}          & 58.59 & \textbf{79.75} & \textbf{62.98}\\
    \bottomrule
    \end{tabular}
}
    \vspace{0.5em}
    \caption{\textbf{Quantitative comparisons to large scale models.} On the test split of Pororo-SV and Flintstones-SV dataset. All models are implemented on top of the large pretrained transformer (\ie, StoryDALL-E~\cite{storydalle} pretrained on 14 million and MEGA-StoryDALL-E~\cite{storydalle} pretrained on 15 million from Conceptual Caption dataset~\cite{sharma2018conceptual} ). In~\cite{storydalle}, the `prompt-tune' update 30\% of model parameters compared to full `fine-tune'.  Without pretraining, we use a much small model compared to the proir arts, \ie, 96.6M (\method) \vs 1.3B (StoryDALL-E).} 
\label{tab:storycont_quan}
\end{table}

\subsection{Human Preference Study}

\begin{table}[t]
\centering
\resizebox{1.0\linewidth}{!}{
    \begin{tabular}{clccc}
    \toprule
    Resolution&Attribute      & VLC-SG~\cite{maharana2021integrating}    & Tie       & Ours    \\
    \midrule
    &Visual Quality   & 27.8\%     & 8.6\%     & \textbf{63.6}\% \\
    &Temproal Consistency      & 24.3\%     & 8.3\%     & \textbf{59.0}\% \\
    \multirow{-3}{*}{\shortstack[c]{$64\times64$\\ (Low-Res.)}}&Semantic Relevance        & 31.2\%     & 10.8\%    &\textbf{57.9}\%  \\ 
    \midrule
    &Visual Quality   & 21.9\%     & 1.5\%     & \textbf{76.6}\% \\
    &Temporal Consistency      & 21.8\%     & 2.5\%     & \textbf{75.7}\% \\
    \multirow{-3}{*}{\shortstack[c]{$128\times128$\\ (High-Res.)}}&Semantic Relevance        & 23.6\%     & 1.7\%    &\textbf{74.6}\%  \\ 
    \bottomrule
    \vspace{0.1em}
    \end{tabular}
}
    \resizebox{1.0\linewidth}{!}{
    \begin{tabular}{lccc}
    \toprule
        & \method w/o Mem. & Tie & \method \\
    \midrule
    Temporal Consistency        &  34.5\% & 4.3\% & \textbf{61.2}\% \\
    \bottomrule
    \end{tabular}
    }
    \vspace{0.5em}
    \caption{\textbf{Human preference studies.} (Top) With 100 judges in the Amazon Mechanical Turk, on Pororo-SV test split dataset. Win (\%) refers to the \% times one model is preferred over the others. `Tie' refers to the same. Ours are clearly preferred over the VLC-SG (refers to VLC-StoryGAN)~\cite{maharana2021integrating}. 
    (Bottom) For the model using our memory (Mem.) module, contrasting results with and without the memory module that considers the contexts spread across the sentences, our method significantly improves human preference in temporal consistency.}
    \vspace{-1em}
\label{tab:human_eval}
\end{table}

We conduct a larger scale (than the prior arts) human survey using the Amazon Mechanical Turk platform to qualitatively compare the generated image sequence by our method over VLC-StoryGAN ~\cite{maharana2021integrating} on three criteria, \ie, visual quality, temporal consistency, and semantic relevancy between generated images and descriptions, following~\cite{li2019storygan,maharana2021integrating,maharana2021improving,song2020character}.
Unlike the previous arts that employ 2 to 9 human judges~\cite{li2019storygan,maharana2021integrating,maharana2021improving,song2020character,vpcsv}, we employ 100 judges for statistically more reliable results.

As shown in the top table in Tab.~\ref{tab:human_eval}, it is clear that human subjects prefer the generated images by our model over those from VLC-StoryGAN ~\cite{maharana2021integrating} on winning ratio(\%). 
We discuss more details about this study in the supplementary material.

\vspace{-0.5em}
\paragraph{Human preference for memory usage.} We further investigate the benefit of using the proposed memory module (Sec.~\ref{sec:method:awm}) by the human study with 100 annotators in Amazon Mechanical Turk.
For the comparative analysis between \method and \method-w/o-Memory, we use 30 randomly sampled image sequences on Pororo-SV test set to conduct A/B test following~\cite{song2020character}. 
Particularly, the annotators need to consider the temporal consistency with semantic relevance between generated image sequence and captions. 
As depicted in the bottom table in Tab.~\ref{tab:human_eval}, we observe that 61.2\% of annotators prefer our \method with memory module over the one without it. 
We can conclude that the memory module has a great deal in creating temporally consistent image sequence with respect to semantic relevancy.

\subsection{Qualitative Analysis}
\label{sec:experiment:qual}
\begin{figure}[t]
    \centering
    \includegraphics[width=0.95\columnwidth]{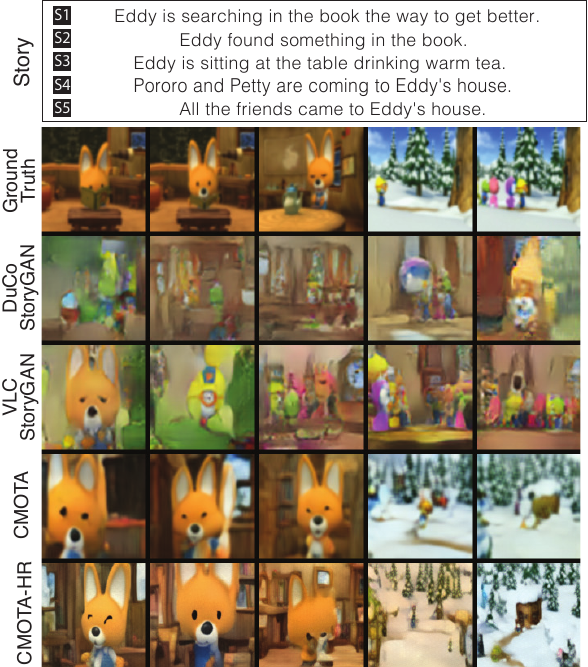}
    \caption{\textbf{Qualitative results compared to prior arts.} On the Pororo-SV's test split. We compare our \method to the prior arts including DuCo-StoryGAN and VLC-StoryGAN. Ours generates a semantically more plausible and temporally coherent image sequence compared to the prior arts. All images except \method-HR (128$\times$128) are generated with resolution of 64$\times$64 for comparison.}
    \vspace{-0.5em}
    \label{fig:quali_main}
\end{figure}

We now qualitatively analyze the quality of generated image sequences in Fig.~\ref{fig:quali_main} on the Pororo-SV dataset's test split.
The top row shows the image sequence of ground truth, the two rows (2-3) contain prior works~\cite{li2019storygan,maharana2021improving} and the final row is the image sequence generated by \method and its high resolution version (\method-HR).
\method generates a semantically more plausible image sequence with better visual quality, compared to prior works~\cite{maharana2021improving, maharana2021integrating}.
Particularly, the \method-HR demonstrates the ability to generate visual details more effectively even as the size of the image increases, while maintaining similar semantics.

\begin{figure}[t]
    \centering
    \includegraphics[width=0.95\columnwidth]{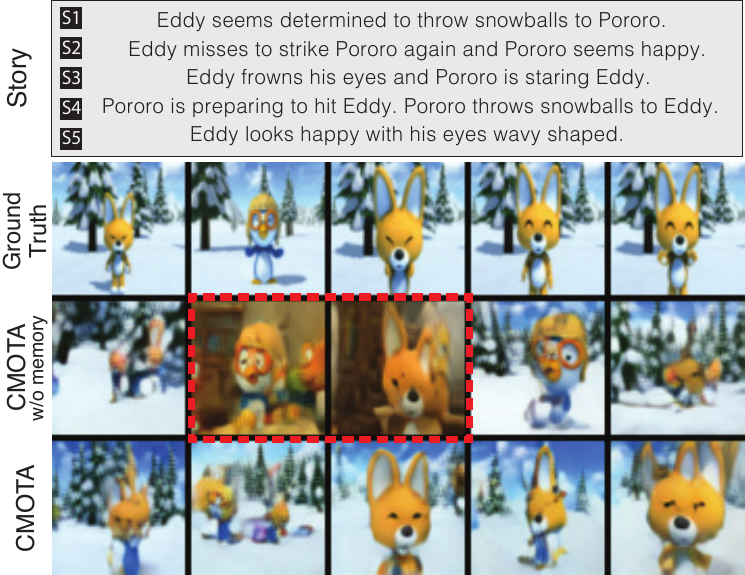}
    \caption{\textbf{Benefit of memory module in encoding background context.} Our \method generates both semantically and visually plausible image sequences. Without memory module, inconsistent images (\ie, abrupt background change) are generated as indicated by the red dotted box.}
    \vspace{-1.5em}
    \label{fig:quali_cohe}
\end{figure}

Furthermore, we qualitatively investigate the advantage of using memory architecture and summarize the results in Fig.~\ref{fig:quali_cohe}.
As shown in the figure, the proposed memory architecture generates a semantically more plausible image sequence with proper context, \eg, background; the \method without memory fails to capture proper background context (shown in dotted red box) since a single sentence could be interpreted in many ways. 
In contrast, \method (with the context memory) generates an image sequence with plausible background with characters preserved without an explicit character-centric model.

\section{Conclusion}
We propose to better encode semantic context, \eg, plausible background and characters, for story visualization using a new memory architecture in a multi-modal bi-directional Transformer.
We further propose an online text augmentation training scheme to generate pseudo-text descriptions as an intermediate supervision for addressing linguistic diversity in the texts at inference.
The proposed method generates a temporally coherent and semantically relevant image sequence for each sentence in the given text paragraph.

The proposed method outperforms prior works by a large margin on various metrics on the two popular SV benchmark datasets, and also outperforms some of hyper-scale models in multiple semantic understanding metrics.
Although computationally prohibited and the pre-trained models are not publicly available, it would be intriguing to apply the proposed memory module and the online augmentation scheme to a large model.